\documentclass[letterpaper, 10 pt, conference]{ieeeconf}

\IEEEoverridecommandlockouts                              
\usepackage[dvipdfmx]{graphicx}
\usepackage{amsmath}
\usepackage{amssymb}
\usepackage{siunitx}
\usepackage{booktabs}
\usepackage{multirow}
\usepackage{balance}
\usepackage{cite}
\usepackage{url}
\usepackage{subcaption}
\usepackage{color}
\usepackage{verbatim}
\usepackage{xcolor}
\usepackage{ifthen}

\def\eg{{\it e.g.}}

\def\ie{{\it i.e.}}

\title{\LARGE \bf
Opt-in Camera: Person Identification in Video via\\UWB Localization and Its Application to Opt-in Systems
}

\author{Matthew Ishige$^{1}$, Yasuhiro Yoshimura$^{1}$ and Ryo Yonetani$^{1}$
\thanks{$^{1}$ M.~Ishige, Y.~Yoshimura, and R.~Yonetani are with Cyber~Agent,~Inc.,
        Tokyo,~Japan
        {\tt\small \{ishige\_mashu, yoshimura\_yasuhiro, yonetani\_ryo\}@cyberagent.co.jp}}%
}

\begin{document}

\maketitle
\thispagestyle{empty}
\pagestyle{empty}

\begin{abstract}
    \looseness=-1
    This paper presents \emph{opt-in camera}, a concept of privacy-preserving camera systems capable of recording only specific individuals in a crowd who explicitly consent to be recorded. Our system utilizes a mobile wireless communication tag attached to personal belongings as proof of opt-in and as a means of localizing tag carriers in video footage. Specifically, the on-ground positions of the wireless tag are first tracked over time using the unscented Kalman filter (UKF). The tag trajectory is then matched against visual tracking results for pedestrians found in videos to identify the tag carrier. Technically, we devise a dedicated trajectory matching technique based on constrained linear optimization, as well as a novel calibration technique that handles wireless tag-camera calibration and hyperparameter tuning for the UKF, which mitigates the non-line-of-sight (NLoS) issue in wireless localization.
    We implemented the proposed opt-in camera system using ultra-wideband (UWB) devices and an off-the-shelf webcam. Experimental results demonstrate that our system can perform opt-in recording of individuals in real-time at 10~fps, with reliable identification accuracy in crowds of 8--23 people in a confined space.
\end{abstract}

\section{Introduction}
\looseness=-1
Cameras are prevalent in a variety of robotics applications, such as mobile robots, social robots, and intelligent surveillance systems. Yet, when it comes to the practical deployment of such camera-equipped robots, privacy and personal data protection pose significant challenges. The EU's General Data Protection Regulation (GDPR) mandates that business stakeholders obtain valid consent from individuals when processing their personal data, including visual features that can lead to personal identification, specifically for direct marketing applications. Even if not legally required, obtaining informed consent can often be best practice from an ethical perspective, especially in sensitive settings like hospitals and homes.

\looseness=-1
In this work, we introduce the concept of privacy-preserving camera systems named ``\emph{opt-in camera}'', which records individuals in a crowd only when they have explicitly provided their consent to be recorded (\ie, opt-in). As proof of explicit consent, we leverage small wireless communication tags that can be attached to personal belongings. These tags are used to track their carriers via wireless localization techniques~\cite{UWB_methods_review}, and furthermore, to localize them in recorded videos. For example, consider retail applications. Retailers could provide shopping carts marked ``\emph{OK to recording}'', equipped with wireless tags, alongside regular shopping carts. Opt-in cameras can then track and record only the individuals carrying the OK-carts, without computationally expensive solutions such as personal identification~\cite{person_identification} and re-identification~\cite{person_re-identification}.

\begin{figure*}[!th]
    \vspace{0.6em}
    \centering
    \includegraphics[width=\linewidth]{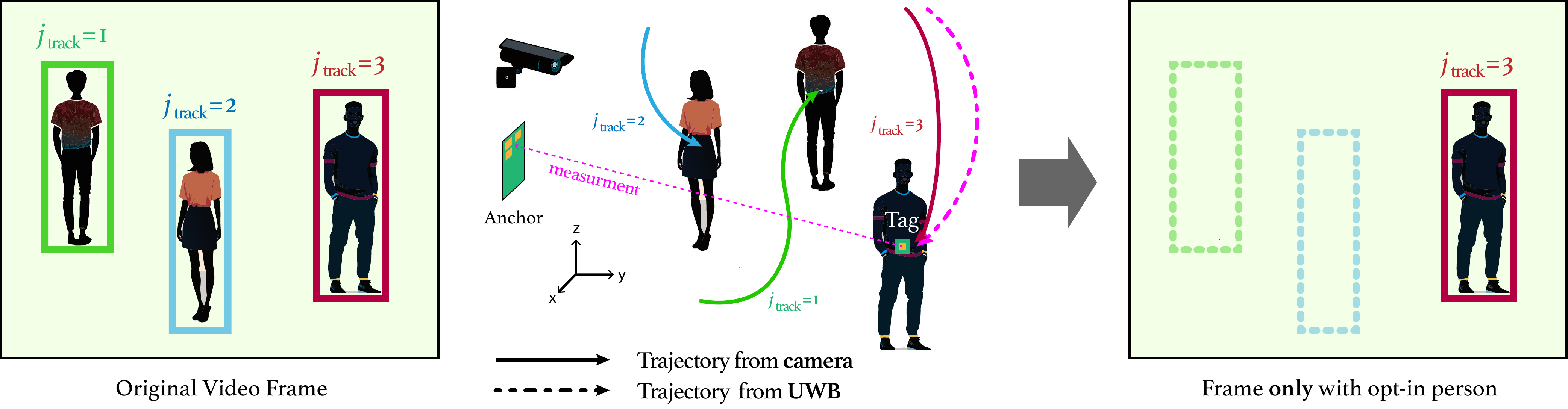}
    \caption{Overview of opt-in camera.}
    \label{fig:overview}
\end{figure*}

As the first implementation of our opt-in camera, we develop a system that effectively integrates a surveillance camera and ultra-wideband (UWB) wireless communication devices. Our system performs opt-in recording as follows (see Fig.~\ref{fig:overview}). First, on-ground position trajectories of UWB tags are tracked using an unscented Kalman filter (UKF). At the same time, a collection of tracklets for every pedestrian in a video is obtained on the same coordinate system via visual tracking.
These tag trajectories and pedestrian tracklets are effectively matched through dedicated constrained linear optimization. To ensure reliable matching even in crowded situations, we develop a novel auto-calibration method that performs wireless tag-camera calibration and hyper-parameter tuning for the UKF from a single pedestrian's walking demonstration, mitigating the non-line-of-sight (NLoS) issue in UWB-based localization.

We conducted extensive experiments to evaluate the proposed opt-in camera system. Experimental results demonstrate that our system can successfully identify people in  crowded scenes (see Fig.~\ref{fig:opt-in_quality}). Our system operates at 10~fps even on consumer-grade laptops (Apple's M3 Max MacBook Pro), indicating its practical usefulness.

\section{Related work}

\subsection{Privacy Protection for Camera Measurement}
Ensuring privacy in visual surveillance has been a critical topic in computer vision. Popular approaches include adding noise to recorded videos to make it difficult to identify individuals~\cite{face_anonymizer,minusface}. Another approach is to use sensors with such low resolution that individuals are difficult to identify to minimize the collection of unnecessary personal information~\cite{low_spatiotemporal_resolution,privacylens}. However, these methods anonymize everyone in the entire image/video, and cannot selectively opt-in or opt-out specific individuals. On the other hand, some prior work aims to identify and track individuals with a wearable cameras~\cite{head_motion_signature,ego_surfing,identify_first-person,actor_observer}. This line of techniques can allow users to search for themselves in videos recorded by others, and perform post-processing such as masking. In contrast to wearable cameras, which have their own privacy concerns, our study proposes leveraging UWB-based position estimation to achieve opt-in privacy protection.

\subsection{Person Identification via Wireless Communication}
\looseness=-1
Various methods have been proposed to identify people or objects in videos using wireless communication technologies. The use of RFID systems is popular~\cite{duan2020enabling}, and more recent work has attempted to integrate deep learning methods in their framework to enable accurate tracking in the 3D space~\cite{rfid_assisted_mot,yin2021robust}. On the other hand, RFID signals are highly susceptible to environmental factors, requiring environment-specific tuning to maintain the accuracy.
Another approach is the use of BLE beacons~\cite{liang2024enhancing}, which can improve localization accuracy by distributing a large collection of beacons in the environment. 
Instead of these wireless localization methods with large errors, we adopted UWB, whose very wide bandwidth enables highly accurate positioning~\cite{al2014comparative}.
Our work only requires a single UWB anchor and a UWB tag for every opt-in individual, and can track the individuals even in crowded scenes thanks to this choice.

\subsection{Robust UWB-based Localization against NLoS}
\looseness=-1
One of the major technical challenges for UWB-based localization is ensuring robustness against non-line-of-sight (NLoS) 
situations~\cite{uwb_nlos_survey} where wireless signals are blocked by environmental obstacles (\eg, furniture and walls) as well as human bodies~\cite{uwb_human_body}. To mitigate errors, it is necessary to first detect the moments when NLoS occurs and then address the resulting errors (see \cite{uwb_nlos_survey} for a recent survey). Statistical/learning-based approaches have been taken for NLoS detection~\cite{nlos_identification_1,nlos_identification_2,nlos_identification_3,nlos_identification_4}. Some work has provided a dedicated benchmark dataset~\cite{uwb_nlos_dataset_cir, uwb_nlos_dataset_rssi}. Unfortunately, learning-based approaches require collecting new training data whenever the environment changes, which significantly increases deployment costs~\cite{nkrow2023transfer}. In contrast, we develop a novel auto-calibration method for UWB-based sensing with minimal data collection and labeling, which requires only the walking demonstration of a single pedestrian holding a UWB tag. As for error mitigation, particle filters (PF) and Kalman filters such as the extended Kalman filter (EKF) are popular choices~\cite{choliz2011comparison}. The former is highly robust and accurate for nonlinear motion, but is computationally intensive and thus not suitable for real-time or edge deployment. The latter is more lightweight, but does not perform as well as PF when tracking targets move rapidly. 
Since the opt-in camera is intended for tracking people walking around indoor facilities and is supposed to run on local edge devices, we opted for a Kalman filter.

\section{Opt-in Camera System}
Figure~\ref{fig:overview} provides an overview of our opt-in camera system. We assume that individuals declare their opt-in for data recording by carrying small UWB tag devices. Our system identifies those tag-carrying, opt-in individuals in camera footage and masks the other people by replacing their regions with background images. As detailed below, this procedure consists of the following four steps: \emph{UWB-based tracking} to estimate on-ground trajectories of the tags (Sec.~\ref{sec:method:uwb}); \emph{camera-based tracking} to obtain a collection of tracklets for each pedestrian found in a video (Sec.~\ref{sec:method:camera_tracking}); \emph{trajectory matching} to identify tag carriers in the video (Sec.~\ref{sec:method:matching}); and \emph{masking} the video so as to contain only the opt-in individuals (Sec.~\ref{sec:method:masking}). Successful identification will require careful calibration of the system in advance, for which we will present a dedicated method in the next section.

\subsection{UWB-based Tracking}
\label{sec:method:uwb}
Positions of the tags are estimated through wireless communication with a fixed UWB anchor device.
This anchor measures the distance and direction of each tag device using time of flight (ToF) and angle of arrival (AoA), first providing tag positions in polar coordinates centered on the anchor: $\mathbf{z}_t = (z_\text{radial}, z_\text{azimuth}, z_\text{elevation})^\top$.
A Bayesian filter then estimates a tag position trajectory $\mathcal{T}^\text{UWB} = \mathbf{p}_{0:T}^\text{UWB}$ from noisy UWB observations, where $\mathbf{p}_t^\text{UWB} = (x, y, z)^\top$ represents the tag position in the global Cartesian coordinate system (the magenta dashed curves in Fig.~\ref{fig:overview}). We adopt the unscented Kalman filter (UKF)~\cite{unscented_kalman_filter}, which we empirically confirmed achieves a good balance between tracking performance and the computational resources required for real-time processing on local devices.
Other filtering techniques, including particle filters, could be used in principle for offline setups.

Specifically, the filter tracks a state $\mathbf{s} = (x, v_x, y, v_y, z)$ of a tag, where $(v_x, v_y)^\top$ represents the on-ground velocities obtained from the UWB measurements $\mathbf{z}_{0:T}$.
The observation model $\mathbf{z} = h(\mathbf{s})$ is a simple coordinate transformation. 
Specifically, the position $(x, y, z)$ is first affinely transformed into the anchor's local Cartesian coordinate system using the anchor's installation parameters, and then converted into polar coordinates.
For the motion model used in the prediction and update steps of the UKF, we adopt a constant velocity model for the $x$ and $y$ components, and a constant position model for the $z$ component, \ie, we assume the $z$ position does not change over time.
When there exist multiple individuals with different opt-in UWB tags, let us describe the trajectory for the $i$-th tag by $\mathcal{T}^\text{UWB}_i$.

\subsection{Camera-based Tracking}
\label{sec:method:camera_tracking}
\looseness=-1
We apply object detection~\cite{yolox} and multi-object tracking (MOT)~\cite{ocsort} algorithms to video footage to obtain pedestrian trajectories in the image space, as shown by the solid cyan, green, and red curves in Fig.~\ref{fig:overview}.
Our object detection algorithm is tailored for humans, enabling detection of bounding boxes for both the entire body and the head region. 
Assuming that the actual width of a human head on average is a constant $w_\text{r}$, the distance of the person from the camera (\ie, depth) can be estimated as $d=f_x\times \frac{w_\text{r}}{w_\text{p}}$, where $f_x$ is the horizontal focal length of the camera and $w_\text{p}$ is the width of the detected head bounding box. By considering the intrinsic and extrinsic parameters of the camera, we can derive the pedestrian tracklets, $\mathcal{T}^\text{cam}_j$ for the $j$-th individual, in the same coordinate system as the tag trajectories. Note that multiple tracklets can correspond to a single individual especially when the environment is crowded and people frequently occlude each other.

\subsection{Trajectory Matching}
\label{sec:method:matching}
\looseness=-1
We match a collection of tag trajectories $\{\mathcal{T}^\text{UWB}_i\}$ against that of pedestrian tracklets, $\{\mathcal{T}^\text{cam}_j\}$ to identify opt-in tag carriers in the video.
We formulate it as a linear assignment problem that assigns a set of tracklets to each tag trajectory.

Let us introduce a binary assignment matrix $X=(x_{i,j})$:
\begin{equation}
    x_{ij} = \begin{cases}
        1 & (\mathcal{T}^\text{cam}_j \sim \mathcal{T}^\text{UWB}_i)   \\
        0 & (\mathcal{T}^\text{cam}_j \nsim \mathcal{T}^\text{UWB}_i),
    \end{cases}
\end{equation}
where $\mathcal{T}^\text{cam}_j \sim \mathcal{T}^\text{UWB}_i$ means the $j$-th pedestrian tracklet is a part of the $i$-th tag trajectory, and $\mathcal{T}^\text{cam}_j \nsim \mathcal{T}^\text{UWB}_i$ otherwise.
To obtain the matching, we solve the following constrained linear optimization problem:
\begin{align}
     & \underset{X}{\max} &  & \sum_{ij} \frac{1}{c_{ij}}\,x_{ij} \label{eq:optimize_objective}                   \\
     & \text{s.t.}        &  & \sum_i x_{ij} \leq 1
    \label{eq:optimize_constrain_one_tag}                                                                         \\
     &                    &  & x_{ij} + x_{ij'} \leq 1 \quad \text{if} \quad |\,T_j \cap T_{j'}| > 0
    \label{eq:optimize_constrain_no_overlap}                                                                      \\
     &                    &  & x_{ij} = 0 \quad \text{if} \quad c_{ij} > c_\text{th}.
    \label{eq:optimize_constrain_cost_threshold}
\end{align}
The cost $c_{ij}$ is a time-averaged distance between tag trajectories and pedestrian tracklets:
\begin{equation}
    c_{ij} = \frac{1}{|\,T_i \cap T_j / T_\text{uc}|} \sum_{t \in T_i \cap T_j / T_\text{uc}} \mathrm{D}(\mathbf{p}_t^{i}, \mathbf{p}_t^{j})
\end{equation}
where $\mathbf{p}_t^i, \mathbf{p}_t^j$ are the positions of $i$-th tag and $j$-th tracklet, and $\mathrm{D}(\cdot, \cdot)$ is the Mahalanobis distance based on the covariance matrix of UKF. $T_i$ and $T_j$ are sets of timestamps in $\mathcal{T}^\text{UWB}_i$ and $\mathcal{T}^\text{cam}_j$, and $T_{uc}$ denotes a set of timestamps where larger uncertainty is found in tag position tracking, which we detect based on if the largest eigenvalue of the covariance matrix of UKF exceeds a threshold $u_\text{th}$.
The constraint~(\ref{eq:optimize_constrain_one_tag}) ensures each pedestrian tracklet is assigned to at most one tag trajectory, while the constraint~(\ref{eq:optimize_constrain_no_overlap}) prevents simultaneous pedestrian tracklets from being assigned to the same tag trajectory.
The constraint~(\ref{eq:optimize_constrain_cost_threshold}) eliminates pairs of tracklets and tag trajectories that are too far apart.

\subsection{Masking}
\label{sec:method:masking}
Once we identify pedestrian tracklets matched to tag trajectories, the post-processing for opt-in is straightforward. First, we prepare a clean background image without any person, simply by taking that image when no pedestrians were present or taking the average of recordings over time. Then we overlay regions (bounding boxes, or semantic segmentation results if computation resources are abundant) of identified, opt-in individuals onto the image.

\section{System Calibration}
Our system has several parameters to be determined in advance, including camera intrinsic/extrinsic parameters that can easily be
tuned through a standard camera calibration as well as other parameters related to UWB sensing that need to be tuned in a dedicated fashion.

\subsection{Extrinsic Calibration between UWB Anchor and Camera}
\label{sec:method:installation_parameter_optimization}
\looseness=-1
Matching UWB tag trajectories with pedestrian tracklets requires extrinsic calibration between the camera and the UWB anchor.
The key parameters here involve the anchor's position $\mathbf{p}^\text{anchor}$, antenna direction $\mathbf{r}^\text{anchor}$ relative to the camera, a tag's height $h_\text{tag}$\footnote{Here we assume the tag's height to be consistent throughout the recording as it is kept in a pocket or attached to its carrier's belongings such as shopping carts/bags.}, and the average head width $w_\text{r}$.
To determine these parameters, we propose a calibration method that involves a person walking around within the camera's field of view while carrying the tag.
Specifically, suppose that the global coordinate system for the camera is defined via a standard camera calibration procedure, and we are given the sequence of UWB data and the corresponding person positions in the video frames.
Using the homography transformation derived from $\mathbf{p}^\text{anchor}$ and $\mathbf{r}^\text{anchor}$, we obtain the tag position $\mathbf{p}^\text{UWB}_t$ in the global coordinate system.
We also estimate a pedestrian tracklet in the same coordinate system with the method in Sec.~\ref{sec:method:camera_tracking} using $w_\text{r}$, assuming the $z$-component (height) of $\mathbf{p}^\text{cam}_t$ is constant at $h_\text{tag}$. We then minimize the discrepancy between $\mathbf{p}^\text{UWB}_t$ and $\mathbf{p}^\text{cam}_t$ over $|T_\text{calib}|$ frames with respect to the calibration parameters:
\begin{align}
    \underset{\substack{\mathbf{p}_\text{anchor}, \ \mathbf{r}_\text{anchor} \\w_\text{r},\ h_\text{tag}}}{\operatorname{arg\,min}}\ \frac{1}{|T_\text{calib}|} \sum_{t \in T_\text{calib}} \bigl|\bigl|\mathbf{p}_t^\text{cam} - \mathbf{p}_t^\text{UWB} \bigr|\bigr|.
    \label{eq:install_param_opt}
\end{align}
We solve this problem with random sample consensus (RANSAC)~\cite{RANSAC} to mitigate NLoS errors.

\subsection{NLoS-aware Parameter Tuning for Kalman Filter}
\label{sec:method:nlos_aware_tuning}
\looseness=-1
Human bodies, including those of tag carriers, can block line-of-sight (LoS) signals between UWB tags and the anchor.
In such NLoS situations, the system relies on signals that have bounced off objects such as walls, causing significant positioning errors.
This problem is especially pronounced in crowded spaces where more people obstruct the direct signal paths.

We address this problem in two steps. First, we learn an NLoS detector $g(\mathbf{f}_t) \in \{0, 1\}$ that identifies NLoS from signal-quality metrics $\mathbf{f}_t$ such as channel impulse response (CIR) and signal-to-noise ratio (SNR).
During calibration, samples identified as outliers by RANSAC (\ie, those exceeding a certain error threshold) are labeled as NLoS.

Second, we switch the UKF's observation noise covariance matrix $R$ between NLoS ($R=R_\text{NLoS}$) and LoS ($R=R_\text{LoS}$) based on the detector $g(\cdot)$:
\begin{align}
    p(\mathbf{s}_t\,|\,\mathbf{z}_t) \  = \  g_\mathbf{\theta} & (\mathbf{f}_t)\ p(\mathbf{s}_t\,|\,\mathbf{z}_t\,;\,R_\text{NLoS}) \nonumber                  \\
                                                               & + \{1 - g_\mathbf{\theta}(\mathbf{f}_t)\}\,p(\mathbf{s}_t\,|\,\mathbf{z}_t\,;\,R_\text{LoS}).
    \label{eq:modified_kalman_filter}
\end{align}
We determine $R_\text{NLoS}$ and $R_\text{LoS}$ (both diagonal matrices) via black-box optimization with CMA-ES~\cite{cmaes} by minimizing:
\begin{align}
    \underset{R_\text{LoS},\,R_\text{NLoS}}{\operatorname{arg\,min}} & \quad \operatorname{tr}(R_\text{LoS}) + \operatorname{tr}(R_\text{NLoS})  \label{eq:tracking_param_opt_objective}                                                                     \\
    \text{s.t.}                                                      & \quad \forall t \in T_\text{calib}, \  \mathbf{p}_t^\text{cam} \in \mathcal{T}^\text{cam}_\text{calib}, \  \mathbf{p}_t^\text{UWB} \in \mathcal{T}^\text{UWB}_\text{calib}, \nonumber \\
                                                                     & \quad \mathrm{D}(\mathbf{p}_t^\text{cam},\ \mathbf{p}_t^\text{UWB}) < d_\text{th}\nonumber
\end{align}
where $\mathrm{D}(\cdot, \cdot)$ denotes Mahalanobis distance.
The tag trajectory $\mathcal{T}^\text{UWB}_\text{calib}$ in the constraint is calculated with (\ref{eq:modified_kalman_filter}) using $R_\text{NLoS}$ and $R_\text{LoS}$.

\section{Evaluation}
We evaluated the tag carrier identification performance of the proposed system from two perspectives:
how it performs as the number of individuals in the surveillance area increases and
how it performs as the number of opt-in individuals grows.

\begin{figure}[!t]
    \centering
    \vspace{0.6em}
    \includegraphics[width=\columnwidth]{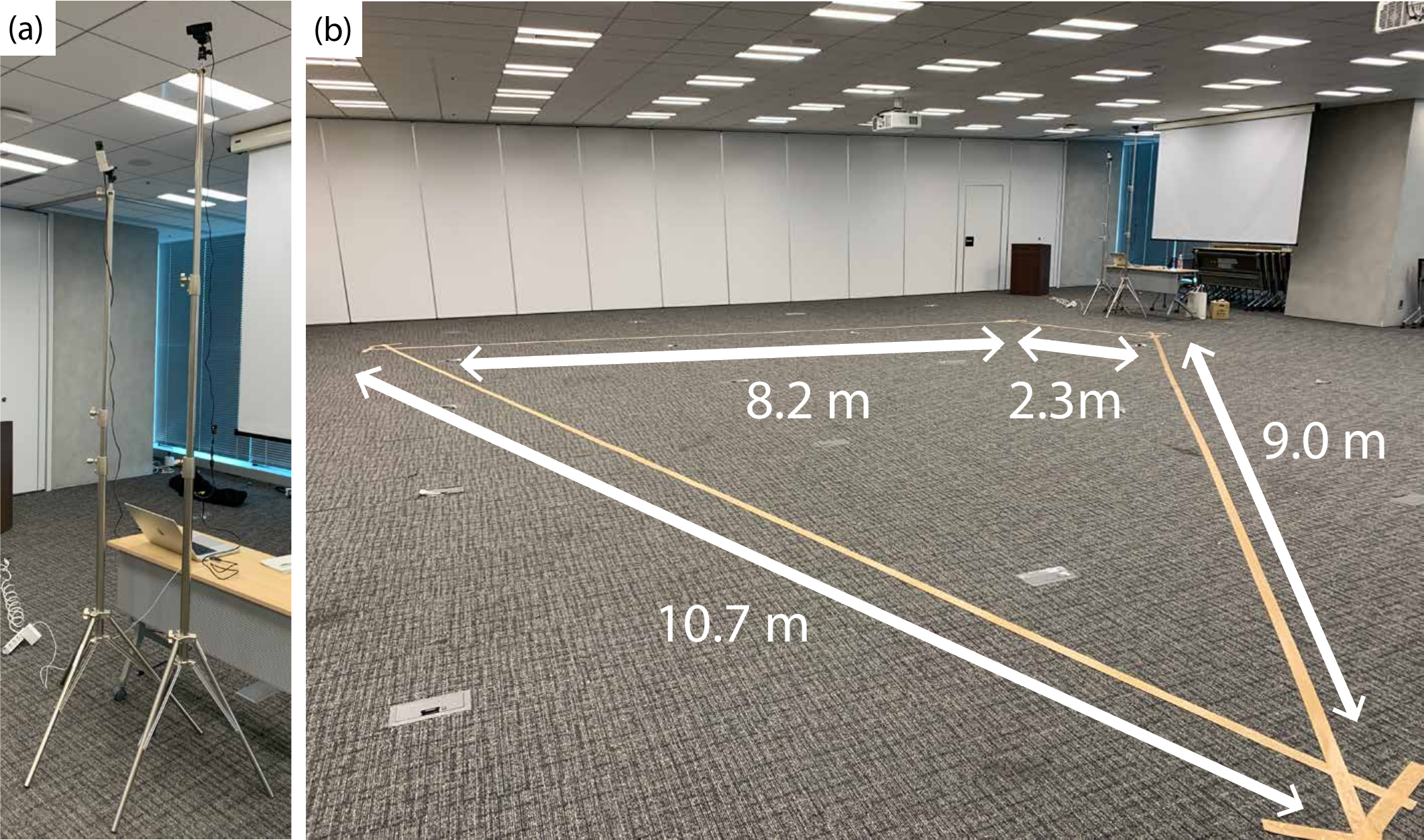}
    \caption{
        (a) A UWB anchor device and a camera fixed on tripods.
        (b) A surveillance area. Subjects randomly walked within this area.
    }
    \vspace{-1.0em}
    \label{fig:experiment_setup}
\end{figure}

\subsection{Implementation Details}
\looseness=-1
We used the object detection model YOLOv9-Wholebody28~(S~ReLU) from PINTO\_model\_zoo~\cite{PINTO-model-zoo}.
For MOT, we adopted OC-SORT~\cite{ocsort} in BoxMOT~\cite{Brostrom2023BoxMOT}.
The camera parameters are determined using OpenCV's standard calibration functions. Specifically, we used a checkerboard pattern to determine the intrinsic parameters, which remain constant as long as we use the same device.
The extrinsic parameters are then determined using ArUco markers. Specifically, we captured a single image of the marker placed on the ground and designated as the origin of the global coordinate system, and then estimated the camera's position and orientation by solving the PnP problem.

We trained the NLoS detector using LightGBM\cite{LightGBM}.
The received signal quality information $\mathbf{f}_t$ was constructed by combining CIR, SNR, RSSI, and an amplitude of a peak in CIR obtained from the anchor antennas.
The process noise for the Kalman filter was calculated using the Q\_discrete\_white\_noise function from filterpy~\cite{filterpy}.
We set the noise for on-ground velocity and height based on the following assumptions:
For velocity, considering movement ranging from stationary to brisk walking, we assumed a maximum change of \SI{2.0}{\meter} during the UWB sampling period of \SI{0.2}{\second}, resulting in a variance of \SI{4.0}{\meter^2}.
For height, since tags are typically carried in pockets, we assumed  a maximum variation of \SI{50}{\centi\meter}, leading to a variance of \SI{0.25}{\meter^2}.
These values are adjusted according to specific application requirements, though we empirically found that performance is not significantly affected by these parameters.

We used $d_\text{th} = 1.0$ in the optimization objective (\ref{eq:tracking_param_opt_objective}).
This value serves as a reference for the association algorithm hyperparameters $c_\text{th}$ and $u_\text{th}$, which were both set to 1.5 to provide a margin above $d_\text{th}$.
As we will demonstrate later, setting these parameters slightly higher than $d_\text{th}$ yields sufficient performance.

\subsection{Experiment Setups}
\subsubsection*{Hardware}
We used LBUA0VG2BP-EVK-P and LBUA2ZZ2DK-EVK from Murata Manufacturing Co., Ltd.\ as our UWB anchor and tags, providing measurements at approximately \qty{5}{Hz} per tag.
Detailed device specifications can be found on their website~\cite{murata2024}.
We employed a C920n webcam from Logicool Co., Ltd.\ for video capture.
All processing was carried out on a consumer-grade Apple MacBook Pro (Apple M3 Max processor, 64~GB memory), where we utilized the built-in GPU to accelerate object detection.
Under these conditions, our implemented system achieved a rate of 10~fps.

\subsubsection*{Field}
\looseness=-1
A trapezoidal surveillance area was marked on the floor of an office room, and subjects were asked to walk randomly within it.
The camera and anchor were set up in front of this area.
Both were mounted on tripods and fixed at a height of above \SI{2.0}{\m} (Fig.~\ref{fig:experiment_setup}).
As long as subjects remained inside the area, their heads were visible in the camera.

\subsubsection*{Baseline Method}
To validate the effectiveness of the proposed method and our calibration approach, we implemented a baseline that instead uses a standard UKF with optimal parameters.
This can be seen as the ideal version of the proposed method, relying on annotated data collected in deployment environments, which is infeasible in actual scenarios.
By comparing our method with this baseline, we will investigate if the proposed method can work comparably well even without such annotated data.
Specifically, we determined the anchor installation parameters through manual calibration, while the remaining parameters were determined through Bayesian optimization using a training dataset (different from the evaluation data).
First, we determined the camera extrinsic parameters using the same procedure as the proposed method --- placing the marker at the origin of the global coordinate system and using OpenCV functions.
Then, to determine the anchor installation parameters $\mathbf{p}^\text{anchor}$ and $\mathbf{r}^\text{anchor}$, we placed ArUco markers at three locations, including the origin.
We sequentially placed a tag device at the center of each marker, collected UWB measurements for \SI{10}{\second} at each location, and calculated the tag position from the medians.
The $\mathbf{p}^\text{anchor}$ and $\mathbf{r}^\text{anchor}$ were determined by minimizing the discrepancy between the calculated tag coordinates and the marker coordinates recognized by the camera.
The other parameters $R$, $w_r$, $c_\text{th}$, and $u_\text{th}$ were determined through Bayesian optimization on the training dataset.
We manually annotated the correct matches and optimized the parameters to maximize accuracy.
Table~\ref{tab:optimized-params} shows the parameters obtained through optimization.
The parameters obtained using the proposed calibration method are close to those obtained through manual optimization in the baseline method.

\begin{table}[t]
    \vspace{0.6em}
    \centering
    \caption{Parameters obtained through optimization in the baseline and the proposed method.}
    \label{tab:optimized-params}
    \begin{tabular}{lcc}
        \toprule
        Parameter & Baseline & Proposed method \\
        \midrule
        $\mathbf{p}_\text{anchor}$   & (0.50,\, -3.21,\, 2.56)      & (0.29,\, -3.29,\, 2.57)      \\
        $\mathbf{r}_\text{anchor}$   & (3.00,\, 0.37,\, 1.55)       & (3.14,\, 0.28,\, 1.56)       \\[0.5ex]
        $w_r$                        & 0.295                        & 0.30                         \\[0.5ex]
        $h_\text{tag}$               & 1.27                         & 1.17                         \\[0.5ex]
        $R_\text{LoS}$               & $\mathrm{diag}(9.0,\, 0.2,\, 0.2)$   & $\mathrm{diag}(9.06,\, 0.10,\, 0.28)$ \\[0.5ex]
        $R_\text{NLoS}$              & ---                          & $\mathrm{diag}(28.99,\, 0.37,\, 0.63)$ \\
        $c_\text{th}$        & 2.5                          & 1.5                          \\[0.5ex]
        $u_\text{th}$        & 2.5                          & 1.5                          \\
        \bottomrule
    \end{tabular}
\end{table}

\begin{table}[t]
    \centering
    \caption{Performance comparison for single opt-in tag identification.}
    \label{tab:single-tag-performance}
    \begin{tabular}{lc}
        \toprule
        Method                    & Recall Rate            \\
        \midrule
        Baseline                  & 0.77 (± 0.36)          \\
        Proposed                  & \textbf{0.86} (± 0.25) \\
        Proposed w/o install opt. & 0.84 (± 0.28)          \\
        Proposed w/o NLoS         & 0.77 (± 0.35)          \\
        Proposed w/o track opt.   & 0.72 (± 0.38)          \\
        \bottomrule
    \end{tabular}
    \vspace{-1.0em}
\end{table}

\subsection{Identification Performance with a Single Opt-in Tag}
\label{sec:experiment:single-optin}
We evaluated the accuracy of the proposed method in identifying an opt-in individual among multiple individuals (from 8 to 23) in the surveillance area, with only one person opting in each scene.
For each scene, we collected one minute of data and split it into six 10-second clips.
We used recall rate as our metric, defined as the proportion of frames where the opt-in person was correctly identified among all frames in which they appeared.
In every scene, the opt-in person was randomly designated and instructed to carry the tag in their right pocket and walk around freely.
Table~\ref{tab:single-tag-performance} shows overall performance across all data.
The proposed method correctly identified the opt-in individual in more than \SI{85}{\percent} of frames on average.
Furthermore, when compared to the baseline method, our approach showed comparable performance (and even outperformed it on average).

Table~\ref{tab:performance-by-people} provides a breakdown by the number of people.
Although performance occasionally dropped depending on who carried the tag, the recall rate generally remained above \SI{80}{\percent} even as the number of people grew and population density increased.

\begin{table}[t]
    \vspace{0.6em}
    \centering
    \caption{Identification performance under different numbers of people.}
    \label{tab:performance-by-people}
    \begin{tabular}{ccc}
        \toprule
        \# of People & Baseline               & Proposed               \\
        \midrule
        8            & \textbf{0.98} (± 0.03) & \textbf{0.98} (± 0.03) \\
        10           & \textbf{0.94} (± 0.04) & \textbf{0.94} (± 0.04) \\
        11           & 0.49 (± 0.49)          & \textbf{0.98} (± 0.01) \\
        12           & \textbf{0.92} (± 0.09) & \textbf{0.92} (± 0.08) \\
        13           & \textbf{0.93} (± 0.06) & \textbf{0.93} (± 0.06) \\
        14           & 0.83 (± 0.10)          & \textbf{0.84} (± 0.10) \\
        15           & 0.76 (± 0.35)          & \textbf{0.91} (± 0.07) \\
        16           & 0.49 (± 0.49)          & \textbf{0.86} (± 0.25) \\
        17           & \textbf{0.56} (± 0.43) & \textbf{0.56} (± 0.43) \\
        18           & \textbf{0.97} (± 0.02) & \textbf{0.97} (± 0.02) \\
        19           & \textbf{0.63} (± 0.44) & \textbf{0.63} (± 0.44) \\
        20           & 0.98 (± 0.02)          & \textbf{0.99} (± 0.01) \\
        21           & 0.61 (± 0.44)          & \textbf{0.83} (± 0.18) \\
        22           & \textbf{0.66} (± 0.40) & \textbf{0.66} (± 0.40) \\
        23           & 0.74 (± 0.34)          & \textbf{0.86} (± 0.10) \\
        \bottomrule
    \end{tabular}
\end{table}

\begin{table}[t]
    \centering
    \caption{Identification performance under different numbers of opt-in tags.}
    \label{tab:performance-by-optin}
    \begin{tabular}{cccccc}
        \toprule
        \# of Tags & 1        & 2        & 3        & 4        & 5        \\
        \midrule
        \multirow{2}{*}{Recall}
                   & 0.98     & 0.81     & 0.83     & 0.79     & 0.89     \\
                   & (± 0.03) & (± 0.29) & (± 0.31) & (± 0.33) & (± 0.19) \\
        \bottomrule
    \end{tabular}
    \vspace{-0.8em}
\end{table}

Additionally, Table~\ref{tab:single-tag-performance} presents ablation study results for our proposed method.
``Proposed w/o install opt.'' omitted the RANSAC-based installation parameter optimization (Sec.~\ref{sec:method:installation_parameter_optimization}) and used parameters obtained through the manual calibration in the baseline method.
``Proposed w/o NLoS'' excluded the NLoS detector (Sec.~\ref{sec:method:nlos_aware_tuning}) and used a single observation error matrix $R$, which is optimized for all scenarios.
``Proposed w/o track opt.'' omitted the tracking parameter optimization (Sec.~\ref{sec:method:nlos_aware_tuning}) and manually designed $R_\text{LoS}$ and $R_\text{NLoS}$ based on certain assumptions.
Specifically, we designed the diagonal matrices assuming distance errors of \SI{0.5}{m} (LoS) and \SI{5.0}{m} (NLoS), and angle errors of \SI{10}{\degree} (LoS) and \SI{45}{\degree} (NLoS).
The results show that each component contributes to performance, with tracking parameter optimization exerting the most significant impact.
This is reasonable, given that the observation error matrix strongly affects Kalman filter performance.
Notably, the variant without the NLoS detector indicates that switching between LoS and NLoS error matrices is more effective than optimizing a single error matrix.

\subsection{Identification Performance with Multiple Opt-in Tags}
\label{sec:experiment:multiple-optin}
We examined how identification performance changes with an increasing number of opt-in individuals.
We fixed the total number of people in the surveillance area at 8 but varied the number of tag carriers from 1 to 5.
Data were recorded for 1 minute under each condition and then divided into six 10-second segments for evaluation.
Table~\ref{tab:performance-by-optin} shows the results.
Overall, identification performance remained stable even as the number of tags increased.

\subsection{Sensitivity to Hyperparameters}
\label{sec:experiment:sensitivity-to-hyperparameters}
\begin{figure}[t]
    \centering
    \vspace{0.6em}
    \includegraphics[width=.95\columnwidth]{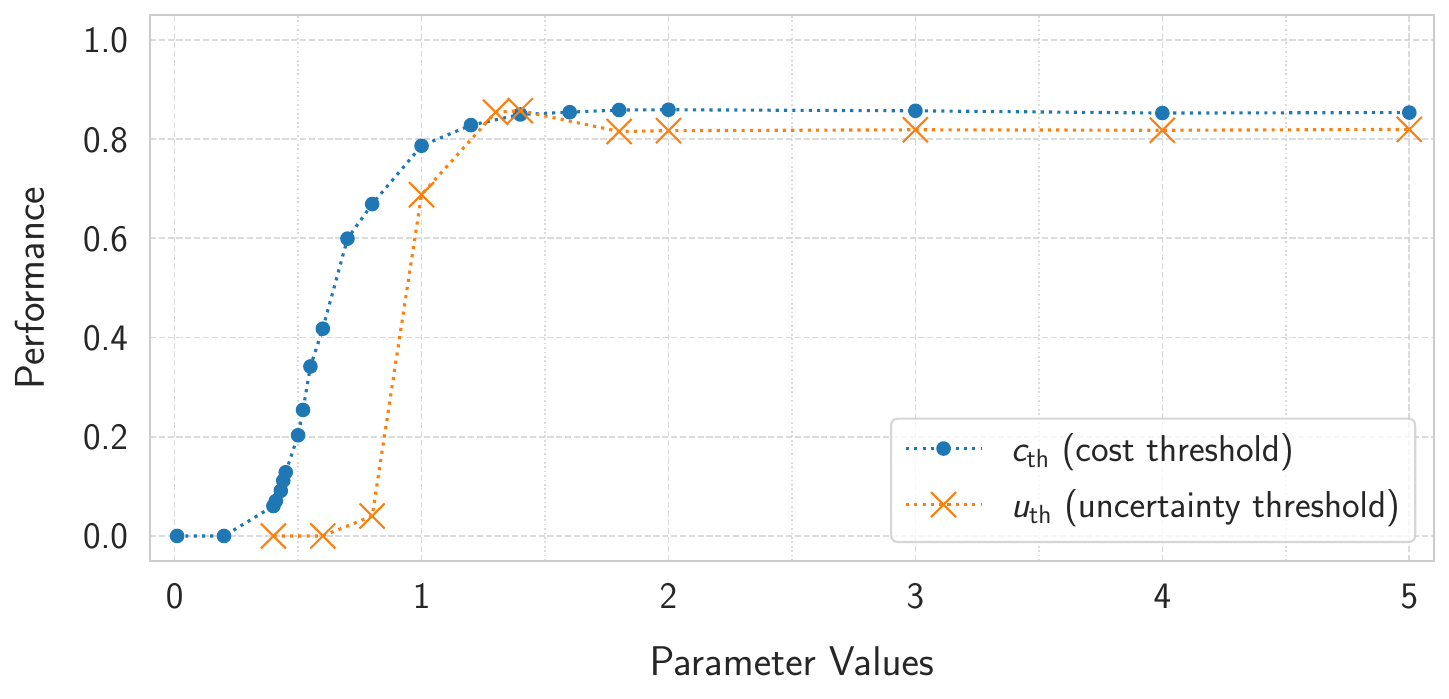}
    \caption{System sensitivity to hyperparameters in the matching algorithm (i.e., $c_\text{th}$ and $u_\text{th}$).}
    \label{fig:parameters_sensitivity}
\end{figure}

\looseness=-1
Most parameters in the proposed method are determined through the optimizations in the calibration.
However, two hyperparameters that significantly affect identification performance remain to be manually set: $c_\text{th}$ and $u_\text{th}$.
Figure~\ref{fig:parameters_sensitivity} shows the results of varying these parameters.
We found that recall rates drop sharply for values below 1.0 but remain consistently high for values above 1.0.
This threshold likely corresponds to our optimization approach, which ensures the Mahalanobis distance between camera footage and UWB-based positioning remains below $d_\text{th} = 1.0$. Consequently, setting $c_\text{th}$ and $u_\text{th}$ slightly above this value yielded satisfactory results.

\begin{table}[t]
    \centering
    \caption{Performance at different time intervals.}
    \label{tab:recall_time_intervals}
    {\footnotesize
        \begin{tabular}{cccccc}
            \toprule
            \SI{5.0}{\second} & \SI{10.0}{\second} & \SI{30.0}{\second} & \SI{60.0}{\second} & \SI{90.0}{\second} & \SI{120.0}{\second} \\
            \midrule
            0.39              & 0.85               & 0.85               & 0.85               & 0.84               & 0.86                \\
            (± 0.46)          & (± 0.28)           & (± 0.26)           & (± 0.27)           & (± 0.29)           & (± 0.25)            \\
            \bottomrule
        \end{tabular}
    }
\end{table}

\subsection{Performance under different calibration durations}
\label{sec:experiment:performance-under-different-calibration-durations}
Since our method requires calibration data, we examined how different calibration data durations affect performance.
Table~\ref{tab:recall_time_intervals} shows the evaluation results using data from Sec.~\ref{sec:experiment:single-optin}, where the duration of calibration data was varied.
Surprisingly, only 10 seconds of calibration data suffices to achieve satisfactory performance.
Since camera extrinsic parameter calibration with OpenCV and ArUco markers takes only a few tens of seconds, the total setup time for our system is about one minute, imposing minimal overhead on installers.

\subsection{Trade-off between precision and recall}
\label{sec:experiment:trade-off-between-precision-and-recall}
\begin{figure}[t]
    \centering
    \vspace{0.6em}
    \includegraphics[width=.95\columnwidth]{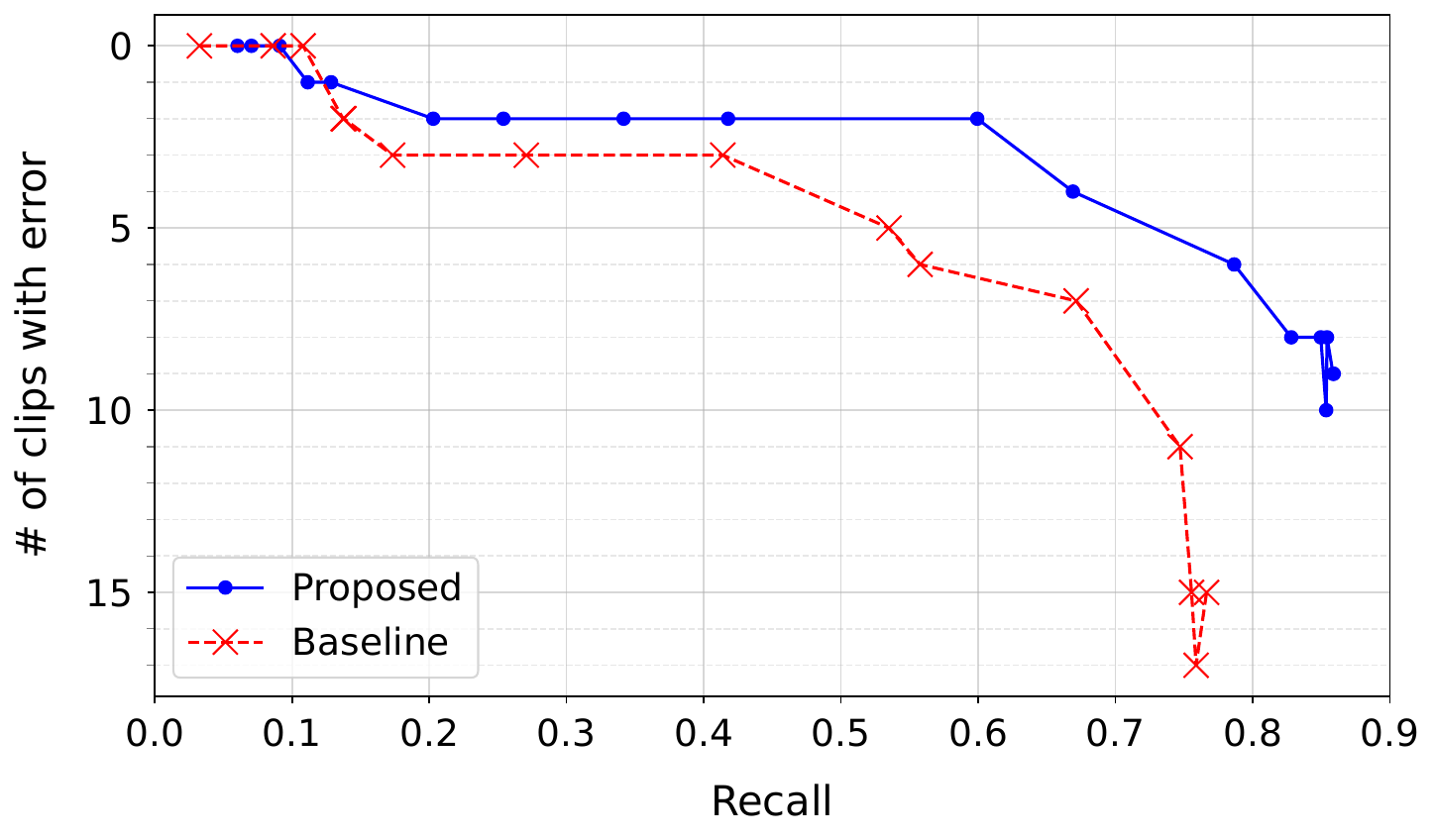}
    \caption{Trade-off between identification errors (inversely related to precision) and recall rate.}
    \label{fig:pr_curve}
\end{figure}

\looseness=-1
In certain applications, misidentifying opt-in individuals can pose serious problems.
In such cases, lowering $c_\text{th}$ reduces the error rate.
Figure~\ref{fig:pr_curve} illustrates this trade-off.
We conducted experiments similar to those in Sec.~\ref{sec:experiment:single-optin}, sweeping $c_\text{th}$ for both the baseline and our proposed method.
A higher $c_\text{th}$ increases recall but raises the likelihood of misidentifications (i.e., lower precision), while a lower $c_\text{th}$ decreases recall yet reduces errors (higher precision).
Although our method achieves a larger area under the curve (AUC) than the baseline, applications highly sensitive to errors may still require significantly higher precision.
Developing methods with better precision-recall trade-offs remains an important challenge for future work; one promising approach may involve incorporating confidence measures into the matching process by analyzing the distribution of the cost matrix.

\subsection{Qualitative Results}
\begin{figure*}[t]
    \centering
    \vspace{0.6em}
    \includegraphics[width=0.93\textwidth]{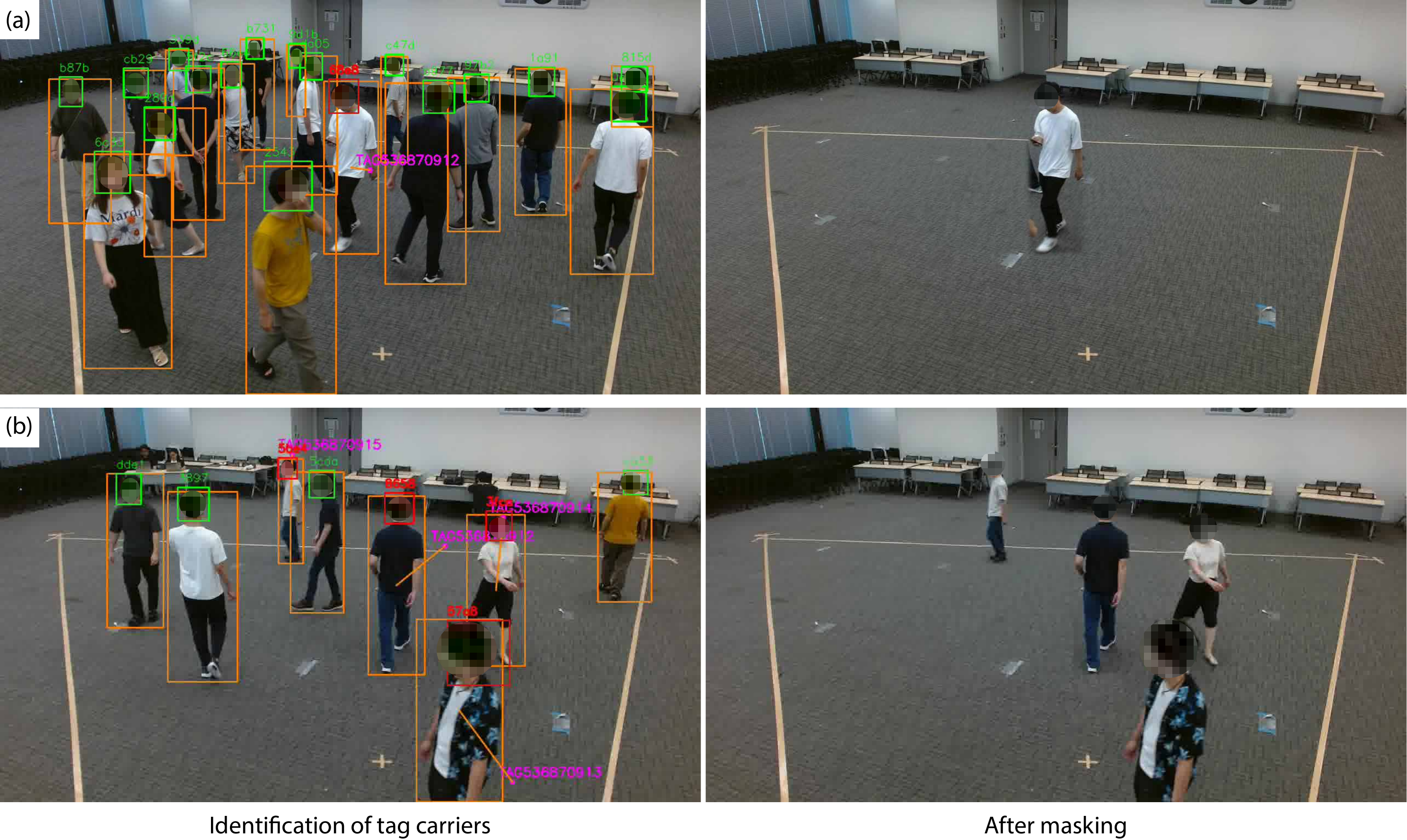}
    \caption{
        Visualization of tag carrier identification and masking results.
        Individuals with red bounding boxes around their heads carried the tags.
        The estimated locations of the tags are shown as magenta points.
        The orange lines show associations between the tags and the corresponding carriers' bounding boxes.
        Note that a mosaic effect is applied to the facial areas to protect the subjects' privacy.
        (a) An example from Sec.~\ref{sec:experiment:single-optin}. One individual carried a tag among 19 others.
        (b) An example from Sec.~\ref{sec:experiment:multiple-optin}. Four individuals carried tags among four others.
    }
    \label{fig:opt-in_quality}
\end{figure*}

Figure~\ref{fig:opt-in_quality} shows examples of opt-in camera outcome.
It demonstrates the feasibility of the proposed opt-in camera system even in crowded space.
We also confirmed the effectiveness of our system in an actual retail store. Please refer to the supplemental video for the demonstration.

\section{Discussions}
\label{sec:discussion}

\subsubsection*{Influence of Object Detection and MOT on Performance}
Our experimental results indicate that the proposed system can robustly track individuals in typical indoor settings, such as retail stores.
The object detector used in our implementation has been trained on a broad range of datasets, facilitating reliable detection under diverse lighting and crowd densities.
However, extremely dense scenarios (e.g., concert venues or sports arenas), settings with extreme lighting conditions, or environments where individuals move abruptly may pose significant challenges for both detection and MOT.
In these specialized contexts, the system may benefit from domain-specific detectors or end-to-end MOT approaches that have been tuned to handle these extreme conditions.

\subsubsection*{Dealing with Moving Cameras}
In this study, we primarily assume static cameras, such as typical surveillance installations.
Under these circumstances, our experiments show that even straightforward masking strategies effectively realize the opt-in functionality.
By contrast, scenarios involving moving cameras (e.g., cameras mounted on robots) are substantially more complex and directly applying our current method is problematic.
To address this challenge, potential strategies include image inpainting through generative models or photometric reconstruction techniques that leverage previously captured frames.
Evaluating and adapting these methods to mobile camera settings remains an important direction for future investigation.

\subsubsection*{Performance Under NLoS Condition}
UWB positioning is commonly considered vulnerable to NLoS conditions.
Nonetheless, our method has been experimentally shown to function reliably even in crowded environments where human bodies themselves create transient NLoS conditions.
The adaptive NLoS detection mechanism suggests that the system may also handle NLoS conditions caused by obstacles other than human bodies.
However, a formal examination of the system's performance under various NLoS scenarios remains a key avenue for future work.

\balance

\subsubsection*{Vulnerability to Malicious Tag Placement}
An additional risk arises from the possibility of maliciously attaching tags to individuals who have not opted in.
To mitigate this threat, it is advisable to enclose tags in conspicuous containers, making covert placement more difficult.
For instance, in a retail context, \emph{opt-in baskets} or \emph{opt-in carts} could be equipped with embedded tags, providing customers with tangible benefits (e.g., discounts) as incentives.
Another potential approach involves shifting from an opt-in to an opt-out paradigm, thereby lessening the advantages that malicious tagging would confer.

\section{Conclusion}
We introduced the concept of an opt-in camera for privacy-preserving sensing, implemented with UWB-based localization.
Consenting individuals carry a UWB tag, which the system tracks and identifies in the video footage.
By masking all non-consenting individuals, the system produces videos that include only opt-in participants.
To address NLoS errors in UWB localization, we proposed a calibration method requiring only a few tens of seconds of walking data during installation.
Our extensive evaluation confirmed the system's effectiveness.
We believe this technology resolves privacy concerns and promotes the practical use of camera-based environmental sensing.

\bibliographystyle{IEEEtran}
\bibliography{IEEEabrv,reference}

\begin{thebibliography}{10}
\providecommand{\url}[1]{#1}
\csname url@rmstyle\endcsname
\providecommand{\newblock}{\relax}
\providecommand{\bibinfo}[2]{#2}
\providecommand\BIBentrySTDinterwordspacing{\spaceskip=0pt\relax}
\providecommand\BIBentryALTinterwordstretchfactor{4}
\providecommand\BIBentryALTinterwordspacing{\spaceskip=\fontdimen2\font plus
\BIBentryALTinterwordstretchfactor\fontdimen3\font minus \fontdimen4\font\relax}
\providecommand\BIBforeignlanguage[2]{{%
\expandafter\ifx\csname l@#1\endcsname\relax
\typeout{** WARNING: IEEEtran.bst: No hyphenation pattern has been}%
\typeout{** loaded for the language `#1'. Using the pattern for}%
\typeout{** the default language instead.}%
\else
\language=\csname l@#1\endcsname
\fi
#2}}

\bibitem{UWB_methods_review}
F.~Mazhar, M.~G. Khan, and B.~S{\"a}llberg, ``Precise indoor positioning using {UWB: A} review of methods, algorithms and implementations,'' \emph{Wireless Personal Communications}, vol.~97, no.~3, pp. 4467--4491, 2017.

\bibitem{person_identification}
J.~Deng, J.~Guo, N.~Xue, and S.~Zafeiriou, ``{ArcFace:} additive angular margin loss for deep face recognition,'' \emph{IEEE Transactions on Pattern Analysis and Machine Intelligence}, vol.~44, no.~10, pp. 5962--5979, 2022.

\bibitem{person_re-identification}
D.~Fu, D.~Chen, J.~Bao, H.~Yang, L.~Yuan, L.~Zhang, H.~Li, and D.~Chen, ``Unsupervised pre-training for person re-identification,'' in \emph{Proc. of the IEEE/CVF Conference on Computer Vision and Pattern Recognition (CVPR)}, Nashville, TN, USA, Jun 2021, pp. 14\,745--14\,754.

\bibitem{face_anonymizer}
Z.~Ren, Y.~J. Lee, and M.~S. Ryoo, ``Learning to anonymize faces for privacy preserving action detection,'' in \emph{Proc. of the European Conference on Computer Vision (ECCV)}, Munich, Germany, Sep 2018, pp. 639--655.

\bibitem{minusface}
Y.~Mi, Z.~Zhong, Y.~Huang, J.~Ji, J.~Xu, J.~Wang, S.~Wang, S.~Ding, and S.~Zhou, ``Privacy-preserving face recognition using trainable feature subtraction,'' in \emph{Proc. of the IEEE/CVF Conference on Computer Vision and Pattern Recognition (CVPR)}, Seattle, WA, USA, Jun 2024, pp. 297--307.

\bibitem{low_spatiotemporal_resolution}
J.~Dai, J.~Wu, B.~Saghafi, J.~Konrad, and P.~Ishwar, ``Towards privacy-preserving activity recognition using extremely low temporal and spatial resolution cameras,'' in \emph{Proc. of the IEEE Conference on Computer Vision and Pattern Recognition Workshops (CVPRW)}, Boston, MA, USA, Jun 2015, pp. 68--76.

\bibitem{privacylens}
C.~Hinojosa, J.~C. Niebles, and H.~Arguello, ``Learning privacy-preserving optics for human pose estimation,'' in \emph{Proc. of the IEEE/CVF International Conference on Computer Vision (ICCV)}, Montreal, QC, Canada, Oct 2021, pp. 2553--2562.

\bibitem{head_motion_signature}
Y.~Poleg, C.~Arora, and S.~Peleg, ``Head motion signatures from egocentric videos,'' in \emph{Asian Conference on Computer Vision (ACCV)}, Singapore, Nov 2015, pp. 315--329.

\bibitem{ego_surfing}
R.~Yonetani, K.~M. Kitani, and Y.~Sato, ``Ego-surfing first-person videos,'' in \emph{Proc. of the IEEE/CVF Conference on Computer Vision and Pattern Recognition (CVPR)}, Boston, MA, USA, Jun 2015, pp. 5445--5454.

\bibitem{identify_first-person}
C.~Fan, J.~Lee, M.~Xu, K.~Kumar~Singh, Y.~Jae~Lee, D.~J. Crandall, and M.~S. Ryoo, ``Identifying first-person camera wearers in third-person videos,'' in \emph{Proc. of the IEEE/CVF Conference on Computer Vision and Pattern Recognition (CVPR)}, Honolulu, HI, USA, Jul 2017, pp. 5125--5133.

\bibitem{actor_observer}
G.~A. Sigurdsson, A.~Gupta, C.~Schmid, A.~Farhadi, and K.~Alahari, ``Actor and observer: Joint modeling of first and third-person videos,'' in \emph{Proc. of the IEEE/CVF Conference on Computer Vision and Pattern Recognition (CVPR)}, Salt Lake City, UT, USA, Jun 2018, pp. 7396--7404.

\bibitem{duan2020enabling}
C.~Duan, W.~Shi, F.~Dang, and X.~Ding, ``Enabling {RFID}-based tracking for multi-objects with visual aids{: A calibration-free solution},'' in \emph{IEEE INFOCOM 2020 - IEEE Conference on Computer Communications}, Toronto, ON, Canada, Jul 2020, pp. 1281--1290.

\bibitem{rfid_assisted_mot}
R.~Song, Z.~Wang, J.~Guo, B.~S. Han, A.~H.~Y. Wong, L.~Sun, and Z.~Lin, ``{RFID}-assisted visual multiple object tracking without using visual appearance and motion,'' in \emph{IEEE International Conference on Image Processing (ICIP)}, Kuala Lumpur, Malaysia, Oct 2023, pp. 2745--2749.

\bibitem{yin2021robust}
J.~Yin, S.~Liao, C.~Duan, X.~Ding, Z.~Yang, and Z.~Yin, ``Robust {RFID}-based multi-object identification and tracking with visual aids,'' in \emph{IEEE International Conference on Sensing, Communication, and Networking (SECON)}, Rome, Italy, Jul 2021, pp. 1--9.

\bibitem{liang2024enhancing}
J.~Liang, S.~Mishra, and C.~Wu, ``Enhancing person identification for smart cities{: Fusion} of video surveillance and wearable device data based on machine learning,'' \emph{IEEE Sensors Journal}, vol.~25, no.~13, pp. 23\,253--23\,261, 2025.

\bibitem{al2014comparative}
M.~A. Al-Ammar, S.~Alhadhrami, A.~Al-Salman, A.~Alarifi, H.~S. Al-Khalifa, A.~Alnafessah, and M.~Alsaleh, ``Comparative survey of indoor positioning technologies, techniques, and algorithms,'' in \emph{International Conference on Cyberworlds}, Santander, Spain, Oct 2014, pp. 245--252.

\bibitem{uwb_nlos_survey}
F.~Wang, H.~Tang, and J.~Chen, ``Survey on {NLOS} identification and error mitigation for {UWB} indoor positioning,'' \emph{Electronics}, vol.~12, no.~7, p. 1678, 2023.

\bibitem{uwb_human_body}
A.~G. Ferreira, D.~Fernandes, S.~Branco, A.~P. Catarino, and J.~L. Monteiro, ``Feature selection for real-time {NLOS} identification and mitigation for body-mounted {UWB} transceivers,'' \emph{IEEE Transactions on Instrumentation and Measurement}, vol.~70, pp. 1--10, 2021.

\bibitem{nlos_identification_1}
F.~Che, Q.~Z. Ahmed, J.~Fontaine, B.~Van~Herbruggen, A.~Shahid, E.~De~Poorter, and P.~I. Lazaridis, ``Feature-based generalized gaussian distribution method for {NLoS} detection in ultra-wideband ({UWB}) indoor positioning system,'' \emph{IEEE Sensors Journal}, vol.~22, no.~19, pp. 18\,726--18\,739, 2022.

\bibitem{nlos_identification_2}
H.~Yang, Y.~Wang, C.~K. Seow, M.~Sun, M.~Si, and L.~Huang, ``{UWB} sensor-based indoor {LOS/NLOS} localization with support vector machine learning,'' \emph{IEEE Sensors Journal}, vol.~23, no.~3, pp. 2988--3004, 2023.

\bibitem{nlos_identification_3}
Y.~Pei, R.~Chen, D.~Li, X.~Xiao, and X.~Zheng, ``Fcn-attention{: A} deep learning {UWB} {NLOS/LOS} classification algorithm using fully convolution neural network with self-attention mechanism,'' \emph{Geo-Spatial Information Science}, vol.~27, no.~4, pp. 1162--1181, 2024.

\bibitem{nlos_identification_4}
S.~Sung, H.~Kim, and J.~Jung, ``Accurate indoor positioning for {UWB}-based personal devices using deep learning,'' \emph{IEEE Access}, vol.~11, pp. 20\,095--20\,113, 2023.

\bibitem{uwb_nlos_dataset_cir}
K.~Bregar, A.~Hrovat, and M.~Mohorcic, ``{NLOS} channel detection with multilayer perceptron in low-rate personal area networks for indoor localization accuracy improvement,'' in \emph{Proc. of the 8th Jo{\v{z}}ef Stefan International Postgraduate School Students' Conference}, Ljubljana, Slovenia, 2016, pp. 1--8.

\bibitem{uwb_nlos_dataset_rssi}
V.~Barral, C.~J. Escudero, J.~A. Garc{\'\i}a-Naya, and R.~Maneiro-Catoira, ``{NLOS} identification and mitigation using low-cost {UWB} devices,'' \emph{Sensors}, vol.~19, no.~16, p. 3464, 2019.

\bibitem{nkrow2023transfer}
R.~E. Nkrow, B.~Silva, D.~Boshoff, and G.~P. Hancke, ``Transfer learning-based {NLOS} identification for {UWB} in dynamic obstructed settings,'' \emph{IEEE Transactions on Industrial Informatics}, vol.~20, no.~3, pp. 4839--4849, 2023.

\bibitem{choliz2011comparison}
J.~Ch{\'o}liz, M.~Eguizabal, {\'A}.~Hern{\'a}ndez-Solana, and A.~Valdovinos, ``Comparison of algorithms for {UWB} indoor location and tracking systems,'' in \emph{IEEE Vehicular Technology Conference (VTC Spring)}, Budapest, Hungary, May 2011, pp. 1--5.

\bibitem{unscented_kalman_filter}
E.~A. Wan and R.~Van Der~Merwe, ``The unscented {Kalman} filter for nonlinear estimation,'' in \emph{Proc. of the IEEE Adaptive Systems for Signal Processing, Communications, and Control Symposium}, Lake Louise, AB, Canada, Oct 2000, pp. 153--158.

\bibitem{yolox}
\BIBentryALTinterwordspacing
Z.~Ge, S.~Liu, F.~Wang, Z.~Li, and J.~Sun, ``Exceeding {YOLO} series in 2021,'' arXiv preprint arXiv:2107.08430, Aug 2021, {Accessed: Jul. 31. 2025.} [Online]. Available: \url{https://arxiv.org/abs/2107.08430}
\BIBentrySTDinterwordspacing

\bibitem{ocsort}
J.~Cao, J.~Pang, X.~Weng, R.~Khirodkar, and K.~Kitani, ``Observation-centric {SORT: Rethinking SORT} for robust multi-object tracking,'' in \emph{Proc. of the IEEE/CVF Conference on Computer Vision and Pattern Recognition (CVPR)}, Vancouver, BC, Canada, Jun 2023, pp. 9686--9696.

\bibitem{RANSAC}
M.~A. Fischler and R.~C. Bolles, ``Random sample consensus{: a} paradigm for model fitting with applications to image analysis and automated cartography,'' \emph{Communications of the ACM}, vol.~24, no.~6, pp. 381--395, 1981.

\bibitem{cmaes}
\BIBentryALTinterwordspacing
M.~Nomura and M.~Shibata, ``cmaes: A simple yet practical python library for {CMA-ES},'' arXiv preprint arXiv:2402.01373, Oct 2024, {Accessed: Jul. 31. 2025.} [Online]. Available: \url{https://arxiv.org/abs/2402.01373}
\BIBentrySTDinterwordspacing

\bibitem{PINTO-model-zoo}
\BIBentryALTinterwordspacing
PINTO0309, ``{PINTO}\_model\_zoo,'' GitHub repository, {Accessed: Jul. 31. 2025.} [Online]. Available: \url{https://github.com/PINTO0309/PINTO_model_zoo}
\BIBentrySTDinterwordspacing

\bibitem{Brostrom2023BoxMOT}
\BIBentryALTinterwordspacing
M.~Broström, ``{BoxMOT:} pluggable {SOTA} tracking modules for object detection, segmentation and pose estimation models,'' GitHub repository, {Accessed: Jul. 31. 2025.} [Online]. Available: \url{https://github.com/mikel-brostrom/boxmot}
\BIBentrySTDinterwordspacing

\bibitem{LightGBM}
G.~Ke, Q.~Meng, T.~Finley, T.~Wang, W.~Chen, W.~Ma, Q.~Ye, and T.-Y. Liu, ``Lightgbm{: a highly efficient gradient boosting decision tree},'' in \emph{Proc. of the International Conference on Neural Information Processing Systems (NeurIPS)}, Long Beach, CA, USA, Dec 2017, pp. 3149--3157.

\bibitem{filterpy}
\BIBentryALTinterwordspacing
R.~Labbe, ``{FilterPy: Kalman} filters and other optimal and non-optimal estimation filters in {Python},'' GitHub repository, {Accessed: Jul. 31. 2025.} [Online]. Available: \url{https://github.com/rlabbe/filterpy}
\BIBentrySTDinterwordspacing

\bibitem{murata2024}
\BIBentryALTinterwordspacing
{Murata Manufacturing Co., Ltd.}, ``my {Murata},'' Website, {Accessed: Jul. 31. 2025.} [Online]. Available: \url{https://my.murata.com/ja/home}
\BIBentrySTDinterwordspacing

\end{thebibliography}

\end{document}